%% file: arxiv.tex
\documentclass[10pt,twocolumn,letterpaper]{article}

\usepackage[pagenumbers]{cvpr} % To force page numbers, e.g. for an arXiv version
\usepackage{fontawesome5}
\input{preamble}
\definecolor{cvprblue}{rgb}{0.21,0.49,0.74}
\usepackage[pagebackref,breaklinks,colorlinks,allcolors=cvprblue]{hyperref}
\usepackage{pifont}      % 用于 \ding{51} 显示 ✓
\usepackage{xcolor}      % 支持颜色
\definecolor{orange}{RGB}{255,165,0}
\definecolor{pink}{RGB}{255,192,203}
\def\paperID{10397} % *** Enter the Paper ID here
\def\confName{CVPR}
\def\confYear{2026}

\title{SwiftVLA: Unlocking Spatiotemporal Dynamics for Lightweight VLA\\ Models at Minimal Overhead}

\author{
Chaojun Ni\textsuperscript{\rm 1,2}\footnotemark[1] ~~
Cheng Chen\textsuperscript{\rm 3}\footnotemark[1] ~~
Xiaofeng Wang\textsuperscript{\rm 1,4}\footnotemark[1] ~~
Zheng Zhu\textsuperscript{\rm 1}\footnotemark[1]~\footnotemark[2]  ~~
Wenzhao Zheng\textsuperscript{\rm 4} \\
Boyuan Wang\textsuperscript{\rm 1} ~~
Tianrun Chen\textsuperscript{\rm 3}\footnotemark[2] ~~
Guosheng Zhao\textsuperscript{\rm 1} ~~
Haoyun Li\textsuperscript{\rm 1} ~~
Zhehao Dong\textsuperscript{\rm 1,2} \\
Qiang Zhang\textsuperscript{\rm 5} ~~
Yun Ye\textsuperscript{\rm 1} ~~
Yang Wang\textsuperscript{\rm 1} ~~
Guan Huang\textsuperscript{\rm 1} ~~
Wenjun Mei\textsuperscript{\rm 2}\footnotemark[2] \\
\textsuperscript{\rm 1}GigaAI ~~
\textsuperscript{\rm 2}Peking University ~~
\textsuperscript{\rm 3}Moxin (Huzhou) Technology Co., Ltd. \\
\textsuperscript{\rm 4}Tsinghua University ~~
\textsuperscript{\rm 5}X-Humanoid \\
\small{\textbf{Project Page:} \url{https://Swiftvla.github.io}}
}

\begin{document}

\maketitle

\renewcommand{\thefootnote}{\fnsymbol{footnote}}
\footnotetext[1]{
These authors contributed equally to this work. 
}
\footnotetext[2]{The corresponding authors for this research are: zhengzhu@ieee.org, tianrun.chen@kokoni3d.com, mei@pku.edu.cn.}

\input{sec/0_abstract}

\input{sec/1_intro}
\input{sec/2_relatedwork}
\input{sec/3_method}

\input{sec/4_experiments}
\input{sec/5_conclusion}

\input{sec/X_suppl}

% \clearpage
% \input{sec/6_reproducibility}
% % \input{sec/2_formatting}
% % \input{sec/3_finalcopy}
{
    \small
    \bibliographystyle{ieeenat_fullname}
    \bibliography{main}
}


\end{document}

%% file: preamble.tex
\input{math_commands.tex}

\usepackage{graphicx}
\usepackage{subcaption}
\usepackage{float}
\usepackage{multirow}
\usepackage{booktabs}
\usepackage{makecell}
\usepackage[table]{xcolor}
\usepackage{wrapfig}
\usepackage{makecell}
\definecolor{citecolor}{HTML}{0071BC}
\definecolor{linkcolor}{HTML}{ED1C24}
\definecolor{SecondPink}{RGB}{255,231,186}
\definecolor{BestOrange}{RGB}{255,210,226}
\usepackage{booktabs,siunitx,makecell,xcolor}
\definecolor{mygray}{gray}{.93}
\newcommand{\best}[1]{\textbf{#1}}
\newcommand{\secondbest}[1]{\underline{#1}}

%% file: math_commands.tex
\usepackage{amsmath,amsfonts,bm}

\def\eqref#1{equation~\ref{#1}}
\def\1{\bm{1}}

\DeclareMathAlphabet{\mathsfit}{\encodingdefault}{\sfdefault}{m}{sl}
\SetMathAlphabet{\mathsfit}{bold}{\encodingdefault}{\sfdefault}{bx}{n}

%% file: sec/0_abstract.tex
\begin{abstract}
% The ABSTRACT is to be in fully justified italicized text, at the top of the left-hand column, below the author and affiliation information.
% Use the word ``Abstract'' as the title, in 12-point Times, boldface type, centered relative to the column, initially capitalized.
% The abstract is to be in 10-point, single-spaced type.
% Leave two blank lines after the Abstract, then begin the main text.
% Look at previous \confName abstracts to get a feel for style and length.
Vision–Language–Action (VLA) models built on pretrained Vision–Language Models (VLMs) show strong potential but are limited in practicality due to their large parameter counts. To mitigate this issue, using a lightweight VLM has been explored, but it compromises spatiotemporal reasoning. Although some methods suggest that incorporating additional 3D inputs can help, they usually rely on large VLMs to fuse 3D and 2D inputs and still lack temporal understanding. Therefore, we propose SwiftVLA, an architecture that enhances a compact model with 4D understanding while preserving design efficiency. Specifically, our approach features a pretrained 4D visual geometry transformer with a temporal cache that extracts 4D features from 2D images. Then, to enhance the VLM’s ability to exploit both 2D images and 4D features, we introduce \textit{Fusion Tokens}, a set of learnable tokens trained with a future prediction objective to generate unified representations for action generation. Finally, we introduce a mask-and-reconstruct strategy that masks 4D inputs to the VLM and trains the VLA to reconstruct them, enabling the VLM to learn effective 4D representations and allowing the 4D branch to be dropped at inference with minimal performance loss. Experiments in real and simulated environments show that SwiftVLA outperforms lightweight baselines and rivals VLAs up to $7\times$ larger, achieving comparable performance on edge devices while being $18\times$ faster and reducing memory footprint by $12\times$.

\end{abstract}

\begin{figure}[t]
    \centering
    \includegraphics[width=.99\linewidth]{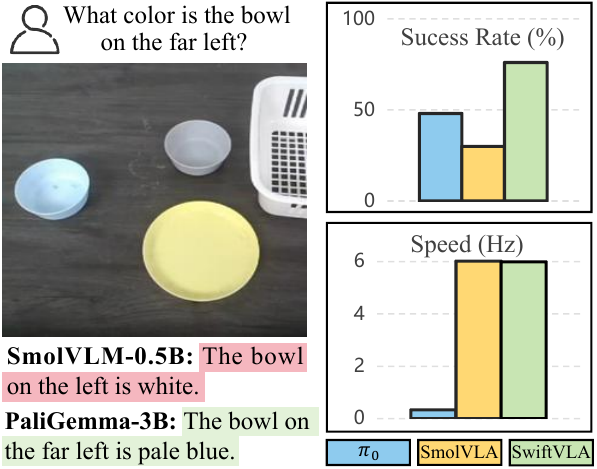} 
\caption{Large VLMs like PaliGemma-3B~\cite{paligemma} excel in spatial reasoning over small VLMs~\cite{smolvlm}, with correct answers in green and incorrect ones in red. This performance advantage allows $\pi_0$~\cite{pi_0} based on it to achieve a higher success rate, despite slower inference speed compared to the SmolVLA~\cite{smolvla} based on a small VLM. However, SwiftVLA enhances spatiotemporal dynamics for small VLA models while preserving the speed advantages. The success rate and speed are tested on the NVIDIA Jetson Orin~\cite{nvidia_jetson_orin}.}
    \label{fig:qa}
    \vspace{-5mm}
\end{figure}

%% file: sec/1_intro.tex
\section{Introduction}
\label{submission}

\begin{figure*}[t]
    \centering
    \includegraphics[width=.99\linewidth]{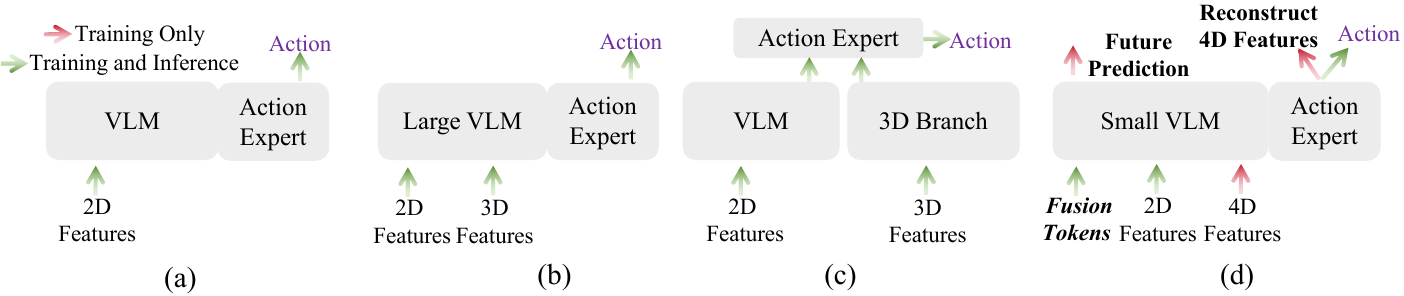} 
    \vspace{-3mm}
% \caption{
% Comparison of methods for integrating spatial and temporal features into VLAs.  
% (a) Using only 2D features as input to the VLM~\cite{pi_0,openvla}, resulting in weak spatiotemporal awareness.  
% (b) Direct fusion approaches that combine spatial features with 2D features inside a large VLM~\cite{bhat20253d,lin2025evo,3dvla}, enhancing spatial reasoning but requiring large models.  
% (c) Decoupled designs that introduce a spatial branch~\cite{geovla,pointvla}, reducing reliance on the VLM but causing substantial parameter overhead.
% (d) Our SwiftVLA leverages a pretrained model~\cite{streamvggt} to extract 4D features and adopts a feature reconstruction objective to compress and align 4D and 2D features. We further introduce \textit{Fusion Tokens} and a future prediction objective to promote tighter integration between different input modalities. The 4D inputs, along with the auxiliary reconstruction and future prediction heads, are used only during training and removed at inference.
% }
\caption{ (a) Using only 2D features as input to the VLM~\cite{pi_0,openvla}, which results in limited spatiotemporal awareness.  
(b) Direct fusion approaches combine spatial and 2D 
features within large VLMs~\cite{bhat20253d,lin2025evo,3dvla}.
(c) Decoupled designs that introduce a dedicated spatial branch~\cite{geovla,pointvla}, causing large parameter overhead.
(d) SwiftVLA leverages a pretrained model~\cite{streamvggt} to extract 4D features and applies a feature reconstruction objective to align 4D and 2D representations. In addition, \textit{Fusion Tokens} and a future prediction objective are introduced to strengthen cross-modal integration. The 4D inputs and auxiliary heads are removed at inference to maintain efficiency.
}
    \label{fig:intro2}
    % \vspace{-2mm}
\end{figure*}

Visual–Language–Action (VLA) models~\cite{pi_0,team2025gigabrain,li2025bridgevla,zheng2025universal,li2025robo,11144414,tangvlascd,li2024automated,liu2025ttf,li2025cogvla,li2025lion,wang2025physiagent,dong2025emma,li2025mimicdreamer,wang2025embodiedreamer,tian2025pdfactor,wang2025flowram,zeng2025futuresightdrive,zheng2023toward,zheng2024odtrack,ye2025vla} represent a new paradigm in robotics, leveraging the representational and reasoning strengths of large, pretrained Vision--Language Models (VLMs)~\cite{paligemma,qwen25,shen2025vlm,jia2024lift3d,chen2025visrl,sarkar2025reasoning,zeng2025futuresightdrive,chang2025scalable} to map natural-language instructions and visual observations directly to actions. Despite their promise, real-world deployment is hindered by a significant obstacle: the massive parameter counts of foundation VLMs induce high inference latency and memory usage, which is prohibitive for real-time control on resource‑constrained robotic platforms.

Therefore, recent studies~\cite{smolvla,tinyvla,minivla} have reduced model capacity by shrinking the size of VLMs or decreasing the number of network layers, enabling deployment on edge devices. However, merely compressing model capacity often weakens reasoning ability, making it difficult to capture the 3D spatial information that is crucial for VLAs to plan precise actions, leading to poor localisation and imprecise trajectories and lowering task success rates. As shown in Fig.~\ref{fig:qa}, smaller VLM models such as SmolVLM-0.5B~\cite{smolvlm} significantly underperform in spatial reasoning tasks, such as answering the question “What color is the bowl on the far left?" compared to larger VLM models~\cite{paligemma}. Therefore, while SmolVLA~\cite{smolvlm}, based on SmolVLM-0.5B~\cite{smolvlm}, exhibits significantly faster inference speeds than $\pi_0$~\cite{pi_0} based on PaliGemma-3B~\cite{paligemma}, its task success rate is notably lower, as complex manipulation tasks often require stronger spatiotemporal reasoning and scene understanding capabilities.

Therefore, recent works~\cite{bhat20253d,lin2025evo,3dvla,geovla,pointvla} have explored integrating 3D and 4D information~\cite{lu2024drivingrecon,lu2025towards,guo2025depth,ding2025nerf,ding2025neural,chen2025think,yao2024waterscenes,drivedreamer4d,recondreamer,recondreamerplus,Recondreamer-rl} to enhance VLAs’ perception of complex environments. However, existing fusion approaches are still suboptimal for lightweight architectures. As shown in Fig.~\ref{fig:intro2}~(b), some methods~\cite{bhat20253d,lin2025evo,3dvla,spatialvla,li2025qdepth,4DVLA} directly fuse 3D features with 2D representations within a large VLM. While this improves spatial awareness compared to Fig.~\ref{fig:intro2}~(a), which uses only 2D input, it has to rely on heavyweight VLMs to handle cross-modal fusion. To mitigate this dependency, other approaches~\cite{geovla,pointvla} (Fig.~\ref{fig:intro2}~(c)) decouple 3D processing from the VLM by introducing an additional branch. However, this design substantially increases parameter overhead, making it unsuitable for compact models. In summary, as shown in Fig.~\ref{fig:intro2}~(a–c), existing approaches still fall short of effectively balancing the lightweight design of VLAs with the practical need for robust and reliable spatiotemporal perception.

In this paper, we present SwiftVLA, a lightweight VLA model built upon a compact VLM~\cite{smolvlm}, which incorporates 4D spatiotemporal information with minimal computational cost. As shown in Fig.~\ref{fig:intro2}~(d), SwiftVLA takes 4D representations as auxiliary inputs and employs a reconstruction objective to learn spatiotemporal dynamics from 4D features, enabling the model to discard them during inference while maintaining performance comparable to full 4D inputs. Meanwhile, \textit{Fusion Tokens} are introduced and supervised by a future prediction objective to promote effective cross-modal fusion. Specifically, SwiftVLA integrates a pretrained 4D visual geometry transformer~\cite{streamvggt} with a temporal cache to convert streaming frames into 4D features incrementally. The cache enables feature reuse across frames and provides temporal context. Meanwhile, because 4D cues are derived directly from standard visual inputs, no additional sensors such as depth cameras or LiDAR are required. For efficient fusion of 2D and 4D features in a compact VLM, we introduce learnable \textit{Fusion Tokens} to unify representations across modalities. Their outputs are supervised by the robot end-effector’s future trajectory to encourage task-relevant learning.  
Finally, we propose a mask-and-reconstruct strategy, where during training, SwiftVLA randomly masks either the 2D or 4D modality with a certain probability and requires the action expert to reconstruct the masked features, which encourages the learning of geometry- and dynamics-aware representations. This enables the model to achieve performance comparable to that with 4D inputs during inference, even without them, minimizing the overhead of 4D inputs while preserving spatiotemporal modeling capability.

% Finally, we propose a mask-and-reconstruct strategy that enables the model to achieve performance comparable to using 4D inputs during inference, even in the absence of 4D inputs. This approach minimizes the overhead of 4D inputs while unlocking spatiotemporal dynamics for lightweight VLA models. Specifically, during training, SwiftVLA randomly masks either the 2D or 4D modality with a certain probability. It then requires the action expert to reconstruct the masked features, thereby promoting the learning of a geometry-aware and dynamics-aware latent representation.
% This supervision enables the \textit{Fusion Tokens} to learn representations that are more effective for VLA's action generation.

% As shown in Fig.~\ref{fig:intro2}~(d), SwiftVLA takes 4D representations as auxiliary inputs and introduces \textit{Fusion Tokens} along with a future prediction objective to enhance cross-modal fusion.
% To further preserve its lightweight design, we propose a mask-and-reconstruct training strategy, which leverages a reconstruction objective to compress and align 4D and 2D features. This strategy allows the model to discard 4D inputs during inference while maintaining performance close to that achieved with full 4D inputs.

\begin{figure*}[t]
    \centering
    \includegraphics[width=0.99\linewidth]{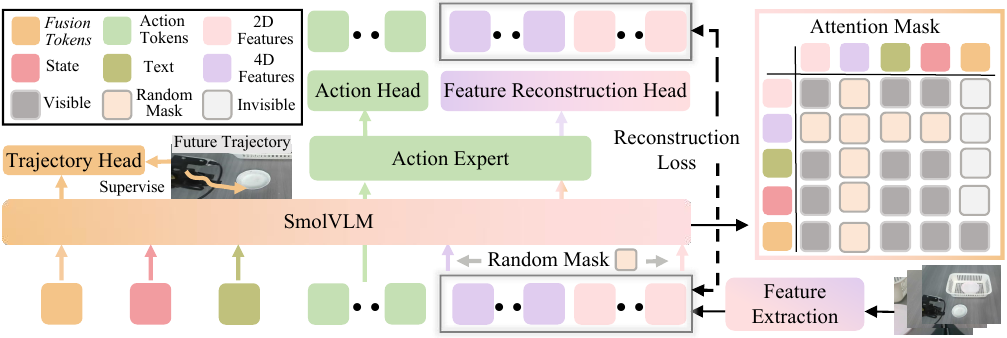} 
    \caption{The pipeline of the SwiftVLA. We first extract 2D and 4D features from input images. A lightweight VLM~\cite{smolvlm} processes 2D and 4D features with \textit{Fusion Tokens} to achieve cross-modal integration. The outputs of the \textit{Fusion Tokens} are supervised by the robot end-effector’s future trajectory. During training, we randomly mask either the 2D or the 4D features, and we require the action expert to reconstruct the masked features while learning to generate actions. We show the attention mask under random masking of the 4D features. In this case, 4D features are excluded from the VLM attention, and the model is required to reconstruct the 4D features from the others.}
    \label{pipeline}
    \vspace{-1mm}
\end{figure*}

We validate SwiftVLA with experiments on simulated and real-world environments. SwiftVLA outperforms lightweight baselines and matches the performance of VLA models up to $7\times$ larger. On edge devices, it maintains comparable performance while running $18\times$ faster than the state-of-the-art baseline $\pi_0$ and reducing memory by $12\times$.
% We validate SwiftVLA with extensive experiments on simulated and real-world environments. SwiftVLA outperforms lightweight baselines and matches the performance of VLA models up to $6\times$ larger. On edge devices, it maintains comparable performance while running $18\times$ faster than the state-of-the-art baseline $\pi_0$ and reducing memory footprint by $10\times$.
% We validate SwiftVLA with extensive experiments on simulated and real-world environemtns. SwiftVLA outperforms lightweight baselines and matches the performance of VLA models up to \(6\times\) larger. On edge devices, it maintains comparable task performance while running \(18\times\) faster than the state-of-the-art baseline \(\pi_0\) and reducing the memory footprint by \(10\times\).

The main contributions of this paper are summarized as follows:

\begin{itemize}
  \item We propose SwiftVLA, a method that integrates 4D spatiotemporal information into a lightweight VLA model at minimal cost. SwiftVLA extracts 4D features and adopts a mask-and-reconstruct training strategy that distills 4D knowledge into the VLA. This enables the model to maintain comparable performance to that with 4D inputs during inference while requiring only 2D inputs. 

    % \item We propose SwiftVLA, integrating 4D spatiotemporal information into a small VLA model at minimal cost. SwiftVLA extracts 4D features and adopts a mask-and-reconstruct strategy that distills 4D knowledge into the VLA. This enables the model to maintain comparable performance to that with 4D inputs during inference while requiring only 2D inputs. 
  
  % We introduce SwiftVLA, a  VLA built on a lightweight VLM that incorporates 4D spatiotemporal cues at minimal cost. Specifically, SwiftVLA employs a 4D visual geometry transformer with a temporal cache to lift 2D images into compact 4D features incrementally, and adopts a mask-and-reconstruct training strategy that distills 4D information into the VLA during training while discarding 4D inputs at inference without sacrificing performance.
  % \item For effective fusion of 2D and 4D features in a lightweight VLM, we introduce a set of learnable \textit{Fusion Tokens} that derive a unified representation across the two modalities. Meanwhile, we guide these tokens toward task-relevant information by supervising their outputs with the robot arm end-effector’s future trajectory.
  \item We fuse 2D and 4D features in a lightweight VLM via learnable \textit{Fusion Tokens}, trained with supervision from the robot arm’s future end-effector trajectory to produce a unified, action-aware representation.
  
  % . Meanwhile, we guide these tokens toward task-relevant information by supervising their outputs with the robot arm end-effector’s future trajectory.  
  % To fuse 2D and 4D features in a lightweight VLM, we introduce learnable \textit{Fusion Tokens} that unify both modalities and are supervised by the robot arm’s future end-effector trajectory.
  
\item Extensive experiments in simulation and on real robots demonstrate that SwiftVLA achieves performance comparable to a baseline that is \(7\times\) larger. On edge devices, it runs \(18\times\) faster and uses \(12\times\) less memory than \(\pi_{0}\).
% Extensive experiments in simulation and on real robots demonstrate that SwiftVLA achieves performance comparable to a baseline that is \(6\times\) larger. On edge devices, it runs \(5\times\) faster and uses \(10\times\) less memory than $\pi_{0}$, significantly reducing end-to-end latency and enabling real-time control.
\end{itemize}

    % \vspace{-2mm}
    % \caption{The pipeline of the SwiftVLA. It uses a pretrained spatiotemporal encoder~\cite{streamvggt} to extract  4D  features from 2D images. Then, a lightweight VLM~\cite{smolvla} jointly processes the 2D and 4D features together with a set of learnable \textit{Fusion Tokens} for cross-modal integration. The outputs of the \textit{Fusion Tokens} are supervised by the robot end‑effector's future trajectory, encouraging representations that are directly useful for action generation. During training, we randomly mask either the 2D or 4D features with a certain probability to enhance modality robustness. We then train the action expert to reconstruct the VLM's 2D and 4D feature inputs while learning to generate actions. }

%% file: sec/2_relatedwork.tex
\section{Related Work}

\subsection{Lightweight VLA Models}
% Recent advances in VLA models~\cite{openvla,pi_0} show superior performance in complex decision-making. These models integrate the backbone of VLMs~\cite{paligemma,deepseek,deepseek2,qwen25} with an action module through end-to-end training, enabling multimodal understanding and motor control.
Recent advances in VLA models integrate VLM backbones with action modules through end-to-end training. OpenVLA~\cite{openvla} introduced a 7B parameter model trained on public datasets~\cite{bridgedata, droid} to generate discrete action tokens. To overcome the limitations of tokenizing actions in continuous control, $\pi_{0}$~\cite{pi_0} uses a diffusion-based decoder to directly generate continuous actions. However, these models~\cite{openvla,pi_0} have a large number of parameters, leading to high training costs and significant inference latency.

To address this, several approaches have shifted toward lighter VLA designs. Based on OpenVLA~\cite{openvla}, MiniVLA~\cite{minivla} replaces the backbone with a smaller model~\cite{qwen25}, thereby reducing the total size to 1B parameters. TinyVLA~\cite{tinyvla} introduces a diffusion policy decoder that directly generates continuous multi-step action sequences to avoid the high latency of autoregressive generation, and employs LoRA~\cite{lora} for parameter-efficient fine-tuning. To further lighten the model, SmolVLA~\cite{smolvla} uses pixel shuffle to limit each frame's tokens and skips a subset of VLM layers, ultimately compressing the parameter count to around 0.5B.

However, to achieve model lightweighting, these methods typically rely on shrinking the backbone parameters, which leads to a degradation in the VLA model's spatial reasoning and fine-grained control capabilities.

\subsection{3D Perception in VLA Models} 
3D perception~\cite{sarkar2025reasoning,wang2025monofusionsparseview4dreconstruction,Zhang2022,shan2021ptt,yu2025cotextor,guo2025depth,wang2025embodiedreamer,9350250,cui20213d,yan2025hemora,liu2023regformer,liu2025mamba4d} is crucial for enhancing robotic manipulation capabilities. Recent studies have attempted to directly incorporate 3D features into the VLM to enhance their geometric awareness, as shown in Fig.~\ref{fig:intro2}~(b). 3D-VLA~\cite{3dvla} extracts spatial embeddings and encodes them into VLM embeddings to improve spatio-temporal reasoning. SpatialVLA~\cite{spatialvla} introduces 3D positional encoding and an adaptive action network into VLMs to improve spatial understanding. Evo-0~\cite{lin2025evo} obtains 3D features by leveraging VGGT~\cite{vggt}, injecting the 3D geometric features into the VLA. However, the domain gap between 2D pixels and 3D geometry is substantial, and directly injecting both into VLMs often requires larger VLMs for better alignment and fusion. Some approaches~\cite{cheng2024spatialrgpt,chen2024spatialvlm,zhou2025vlm4d} also attempt to fine-tune VLMs for spatio-temporal reasoning, but this often relies on massive amounts of temporally annotated data, which are expensive to collect.

Therefore, as shown in Fig.~\ref{fig:intro2}~(c), some methods adopt decoupled designs that introduce a spatial branch. PointVLA~\cite{pointvla} treats point clouds as auxiliary conditioning signals and decouples 3D processing from the 2D vision encoder, enabling the model to leverage geometric cues while preserving the integrity of pretrained 2D representations. GeoVLA~\cite{geovla} adopts parallel branches for multimodal inputs and leverages modality-specific experts to achieve fusion. However, these approaches focus only on 3D information, neglecting temporal dynamics, while increasing memory footprint and inference latency. More recently, 4D-VLA~\cite{4DVLA} incorporates the temporal dimension into VLA modeling by leveraging a history-similarity-based keyframe sampling strategy and generating 3D-aware spatial-visual tokens. While this method enhances spatiotemporal perception, sampling multiple frames introduces additional inference overhead. In contrast, SwiftVLA maintains a lightweight design while injecting 4D cues at lower cost. 

% However, these approaches focus solely on 3D information, neglecting temporal dynamics necessary for action generation, and they add parameters that increase memory footprint and inference latency.

% It uses a 4D visual geometry transformer and a mask-and-reconstruct training scheme to distill 4D information into the VLA during training, enabling 4D inputs to be discarded at inference with minimal performance loss. Meanwhile, we introduce learnable \textit{Fusion Tokens} in the lightweight VLM to unify 2D and 4D representations, and supervise their outputs with the robot arm end-effector’s future trajectory, steering the fusion toward task-relevant spatiotemporal and geometric cues.

% In contrast, SwiftVLA introduces a 4D visual-geometry transformer~\cite{streamvggt} that leverages a temporal cache to incrementally aggregate 4D features from the 2D input, without requiring additional sensors. Meanwhile, we introduce \textit{Fusion Tokens}, guided by the robot’s end-effector trajectory, to fuse 2D features with 4D spatiotemporal features within a small VLM. Finally, the mask-and-reconstruct strategy ensures that 4D feature fusion introduces no additional parameters at inference time.

%% file: sec/3_method.tex
\section{Method}

\begin{figure}[t]
  \centering
  \includegraphics[width=0.99\linewidth]{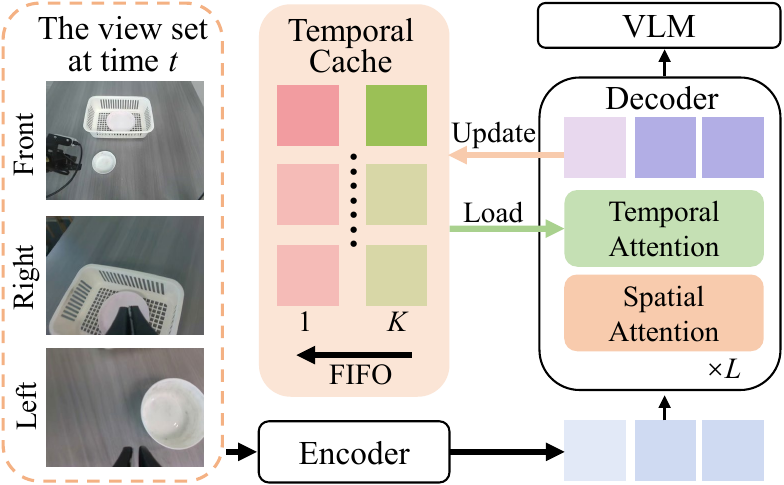}
  \caption{The process of 4D feature extraction. At each step, we sequentially process multi-view observations and load contextual information from the cache for temporal attention. The generated 4D features are updated to the cache and delivered to the VLM.}
  \label{fig:method_3}
  \vspace{-3mm}
\end{figure}

\label{method}
% Existing lightweight VLA models reduce parameters by adopting lightweight VLM backbones~\cite{smolvlm}. However, this parameter reduction limits their spatial reasoning capabilities, causing them to struggle in real-world environments. Although some works~\cite{3dvla,pointvla} attempt to fuse 3D inputs to enhance the spatial perception ability of VLAs, such approaches usually require a large number of parameters to effectively integrate the two different modalities, making them unsuitable for lightweight VLAs. Therefore, we propose SwiftVLA, a framework that incorporates spatio-temporal dynamics into lightweight VLA models, enhancing performance with minimal overhead. 

\subsection{Model architecture}
As shown in Fig.~\ref{pipeline}, SwiftVLA consists of two connected components: a pretrained lightweight VLM~\cite{smolvlm} and an action expert. The VLM  processes the input to extract both 2D and 4D features with a pretrained 4D visual geometry transformer~\cite{streamvggt}. Meanwhile, \textit{Fusion Tokens} are introduced into the VLM to better leverage both 2D and 4D features, supervised by end-effector trajectory prediction. Then, the VLM fuses the 2D and 4D features, \textit{Fusion Tokens}, and other inputs to generate the intermediate hidden states, which serve as the condition for the action expert model to perform action prediction. Additionally, during training, we employ a mask-and-reconstruction strategy in which the 2D or 4D features are masked so that they do not contribute to action generation, and the action expert is required to reconstruct the corresponding features. This strategy encourages the model to exploit cross-modal cues and enables the removal of 4D feature inputs during inference with minimal performance loss, thereby maintaining a lightweight design.

Specifically, at each time step $t$, given the ordered view set \(S = [left, right, front]\), the robot receives a natural language instruction $l$, multi-view observations $o_t = \{o_t^{v}\}_{v \in S}$, and the proprioceptive state $s_t$. From the raw images, we first extract 2D visual features from each view using an image encoder~\cite{zhai2023sigmoid}, obtaining $F_{2D}^{t} = \{F_{2D}^{t,v}\}_{v \in S}$. Then, a pretrained 4D visual geometry transformer~\cite{streamvggt} is applied to derive temporally-enhanced 4D features $F_{4D}^t$. A set of learnable \textit{Fusion Tokens} $Q_f$ are then introduced to interact with $F_{2D}^t$, $F_{4D}^t$, with proprioceptive embeddings $E_s^t$ and language embeddings $E_l^t$ through a lightweight VLM $\mathcal{V}$, resulting in a unified representation:
\begin{equation}
Z_{f}^t = \mathcal{V}(Q_f,  E_s^t, E_l^t, F_{4D}^t, F_{2D}^t).
\label{traj}
\end{equation}
In the fused representation \( Z_f^t \), the portion corresponding to the \textit{Fusion Tokens} is decoded to predict the end-effector trajectory, with explicit supervision from ground-truth trajectories, thereby enabling the intermediate hidden states of $\mathcal{V}$  to learn trajectory-aware cross-modal alignment. In parallel, the intermediate hidden states of $\mathcal{V}$, 
denoted as $\{h_{\mathcal{V}}^{(i)}\}$, 
are employed as hierarchical conditional features for the action expert $\mathcal{A}$.  
The action expert is formulated as a conditional diffusion model. Given a noise sample \(\epsilon\) and conditioned on VLM features \(\{h_{\mathcal{V}}^{(i)}\}\), it produces an action latent: 
\begin{equation}
Z_{\mathcal{A}}^{t} = \mathcal{A}\!\left(\epsilon \,\middle|\, \{h_{\mathcal{V}}^{(i)}\}\right).
\label{eq:action-expert}
\end{equation}
We decode \(Z_{\mathcal{A}}^{t}\) with two complementary heads: the first predicts the diffusion noise for the action, and the second reconstructs the masked feature representations. The reconstruction head is trained with an auxiliary objective to improve cross‑modal alignment and is discarded at inference time, ensuring a lightweight design and fast inference.

\subsection{Incremental 4D Feature Extraction}
\label{4dfeatures}
Recent studies~\cite{bhat20253d,lin2025evo,3dvla,geovla,pointvla} show that integrating 3D information~\cite{vggt,streamvggt,ni2025wonderfree,humandreamerx,wang2025drivegen3d}, such as depth maps and point clouds, can improve the spatial perception of VLA models. Such signals provide geometric cues that help models better reason about distance and object layout. Yet these methods often rely on additional sensors and underutilize VLAs’ temporal modeling capacity, resulting in limited spatio-temporal reasoning.

Therefore, as shown in Fig.~\ref{fig:method_3}, we adopt a pretrained 4D visual geometry transformer~\cite{streamvggt} consisting of an encoder, a decoder, and a temporal cache module, whose weights are kept frozen, that incrementally extracts 4D features from input images, avoiding reliance on extra sensors and leveraging spatial and temporal cues. At each time step $t$, for each view $v \in S$, the observation $o_t^{v}$ is encoded into a feature embedding using the encoder:
\begin{equation}
F_{e}^{t,v} = Encoder(o_t^{v}).
\end{equation}
We sequentially feed the encoded features into a spatiotemporal decoder, which employs spatial and temporal attention to extract 4D representations following the view order $S$. During the temporal attention stage, the current feature embedding interacts with the temporal cache through cross-attention, enabling the integration of temporal context. The temporal cache is then updated iteratively to generate 4D features in a consistent view order. We initialize the temporal cache as $C^{t,0}=C^{t-1}$. For each $k \in \{1,2,3\}$ with the corresponding view $v = S_k$, the decoding process is formulated as:
\begin{equation}
\label{eq:decoder}
(F_{4D}^{t,v},\, C^{t,k})
= \mathrm{Decoder}\!\big(\mathrm{CrossAttn}(F_{e}^{t,v},\, C^{t,k-1})\big).
\end{equation}
After all views have been processed, the temporal cache is updated as $C^{t}=C^{t,3}$. To maintain temporal efficiency, we adopt a First-In-First-Out (FIFO) strategy for the cache, retaining only the most recent $K$ representations of the 4D features. Finally, the visual inputs to the VLM are defined as follows:
\begin{equation}
\begin{aligned}
F_{2D}^{t}=\{F_{2D}^{t,v}\}_{v\in S},\qquad
F_{4D}^{t}=F_{4D}^{t,front}.
\end{aligned}
\end{equation}
The \(F_{4D}^{t,left}\) and \(F_{4D}^{t,right}\) are only used to update the temporal cache to provide a more comprehensive spatiotemporal context and are not provided to the VLM as 4D features, considering the training cost.

\subsection{Fusion Tokens}
Many prior works~\cite{3dvla,spatialvla,pointvla,lin2025evo} attempt to feed 3D information into VLMs so that they can jointly understand 2D and 3D inputs. However, these approaches typically rely on heavyweight VLMs. Lightweight VLMs struggle to develop robust spatial reasoning and cannot effectively fuse multimodal inputs into a coherent, 3D-aware latent space.

To address this problem, we introduce \textit{Fusion Tokens}, a set of learnable tokens that interact with 2D features and 4D spatiotemporal features, and are directly supervised by the end-effector's future trajectory. The keys and values produced by the \textit{Fusion Tokens}, together with those from other tokens, form the conditioning signal \(h_{\mathcal{V}}^{(i)}\) used by the action expert to generate action chunks. Specifically, Fusion Tokens interact with the aggregated multimodal token sequence composed of the 2D features $F_{2D}^{t}$, the 4D features $F_{4D}^{t}$, the language embeddings $E_{l}^{t}$, and the state embeddings $E_{s}^{t}$ through cross-attention within the VLM $\mathcal{V}$, yielding a fused representation $Z_{f}^{t}$, as shown in Eq.~\ref{traj}. The fused representation $Z_{f}^{t}$ serves as the perception output and is optimized via end-effector trajectory prediction:
\begin{equation}
\hat{\tau}_{t} = h_{traj}\!\left(Z_{f}^{t}\right), \qquad
\mathcal{L}_{traj} = \left\lVert \hat{\tau}_{t} - \tau_{t} \right\rVert_{2}^{2},
\end{equation}
where \( h_{traj} \) is a predictor that decodes the trajectory, \( \hat{\tau}_{t} \) is the predicted trajectory, and \( \tau_{t} \) is the ground-truth. Using the end-effector's future trajectory as supervision, \textit{Fusion Tokens} align multimodal features with spatiotemporal semantics, making \( h_{\mathcal{V}}^{(i)} \) more effective for action generation.
% \begin{equation}
% \hat{\tau}_{t} = \mathrm{Head}_{\text{traj}}\!\left(Z_{\mathcal{V}}^{t}\right),
% \end{equation}
% \begin{equation}
% \mathcal{L}_{\text{traj}} = \left\lVert \hat{\tau}_{t} - \tau_{t} \right\rVert_{2}^{2},
% \end{equation}
% \begin{equation}
% \mathcal{L}_{\text{traj}} = \left\lVert \mathrm{Head}_{\text{traj}}\!\left(Z_{f}^{t}\right) - \tau_{t} \right\rVert_{2}^{2}.
% \end{equation}
% where \(\mathrm{Head}_{\text{traj}}\) is a lightweight predictor that decodes the trajectory, \(\hat{\tau}_{t}\) is the predicted trajectory, and \(\tau_{t}\) is the ground-truth trajectory. By grounding the fused embedding in trajectory supervision, the model aligns multimodal perception with actionable spatiotemporal semantics, ensuring that \(Z_{\mathcal{V}}^{t}\) directly benefits downstream robotic control.
% we aggregate the 2D features \(F_{2D}^{t}\), the 4D features \(F_{4D}^{t}\), the language embeddings \(E_{L}\), and the state embeddings \(E_{s}^{t}\) into a multimodal token sequence and process it with the VLM \(\mathcal{V}\) using the \textit{Fusion Tokens} via cross-attention, yielding a fused representation \(Z_{f}^{t}\). The fused representation \(Z_{f}^{t}\) serves as the output embedding of the perception and is optimized via end-effector trajectory prediction:
\subsection{Mask and Reconstruct Strategy}
Although incorporating 4D features with \textit{Fusion Tokens} into VLAs can significantly enhance their spatial reasoning capabilities, the resulting increase in parameters and computational overhead runs counter to the lightweight design principle of VLAs. Therefore, we propose a mask-and-reconstruction strategy that leverages 4D supervision signals during training to build geometry-aware representations, while discarding 4D feature inputs during inference to maintain model efficiency with minimal performance degradation.  Our approach encourages the model to build geometry-aware representations through structured masking and reconstruction, thereby distilling rich spatial and temporal knowledge into the learned features. Next, we present the training and inference procedures in detail.

\begin{table*}[t]
\centering
\setlength{\abovecaptionskip}{0.5em}
\setlength{\tabcolsep}{12pt}
\resizebox{1\linewidth}{!}{
\begin{tabular}{lcccccccc}
\toprule
\multirow{2}{*}{Method} &
\multicolumn{2}{c}{Short-Horizon} &
\multicolumn{2}{c}{Medium-Horizon} &
\multicolumn{2}{c}{Long-Horizon} &
\multicolumn{2}{c}{Average} \\
\cmidrule(lr){2-3}
\cmidrule(lr){4-5}
\cmidrule(lr){6-7}
\cmidrule(lr){8-9}
& SR $\uparrow$ & Length $\downarrow$
& SR $\uparrow$ & Length $\downarrow$
& SR $\uparrow$ & Length $\downarrow$
& SR $\uparrow$ & Length $\downarrow$ \\
\midrule
$\pi_0$~\cite{pi_0} & \secondbest{0.42} & 120 & 0.46 & \secondbest{150} & 0.52 & 187 & 0.47 & 152 \\
GO-1~\cite{bu2025agibot} & 0.40 & 124 & 0.44 & 160 & 0.54 & 190 & 0.46 & 158 \\
TinyVLA~\cite{tinyvla} & 0.08 & 183 & 0.08 & 240 & 0.06 & 236 & 0.07 & 220 \\
SmolVLA~\cite{smolvla} & 0.28 & 152 & 0.32 & 178 & 0.28 & 234 & 0.29 & 188 \\
SmolVLA\textsuperscript{$\dagger$}~\cite{smolvla} & 0.38 & 130 & 0.36 & 165 & 0.34 & 195 & 0.36 & 163\\
\rowcolor{mygray}
SwiftVLA & \best{0.56} & \secondbest{115} & \secondbest{0.48} & 156 & \secondbest{0.56} & \best{180} & \secondbest{0.53} & \secondbest{150} \\
\rowcolor{mygray}
SwiftVLA with 4D input & \best{0.56} & \best{100} & \best{0.50} & \best{145} & \best{0.58} & \secondbest{185} & \best{0.55} & \best{143} \\
\bottomrule
\end{tabular}
}
\caption{Comparison of task success rate and average trajectory length in simulation. The best results are marked in \best{bold}, and the second-best results are \underline{underlined}. {$\dagger$} denotes the model that is pre-trained and fine-tuned using the same configuration as SwiftVLA.}
\label{tab:robotwin_results}
\vspace{-3mm}
\end{table*}

% Although incorporating 4D features with Fusion Tokens can substantially improve the spatial reasoning ability of VLA models, this benefit comes at the cost of a significant increase in parameters and computational overhead. Such complexity poses serious challenges for deployment on resource-constrained robotic platforms.
% To address this limitation, we propose a mask-and-reconstruction strategy that leverages 4D supervision signals only during training. Specifically, our approach encourages the model to build geometry-aware representations through structured masking and reconstruction, thereby distilling rich spatial and temporal knowledge into the learned features.
\noindent
\textbf{Training.} During training, we employ a random masking strategy that, with a certain probability, applies masks to either the 2D or 4D features. Under this setting, the VLA is required to predict actions based on the remaining modalities, while simultaneously reconstructing the masked features. As illustrated in Fig.~\ref{pipeline}, we visualize the masking operation applied to the 4D features during training. The gray and white blocks indicate the fixed visible and invisible tokens, respectively, while the pink blocks represent tokens that undergo random masking, turning originally visible tokens into invisible ones. Meanwhile, the reconstruction losses are defined as follows:
\begin{equation}
\begin{split}
\mathcal{L}_{2D} &= \left\| h_{2D}\!\left(Z_{\mathcal{A}}^{t}\right) - F_{2D}^{t} \right\|_{2}, \\
\mathcal{L}_{4D} &= \left\| h_{4D}\!\left(Z_{\mathcal{A}}^{t}\right) - F_{4D}^{t} \right\|_{2}.
\end{split}
\end{equation}
where $F_t^{2D}$ and $F_t^{4D}$ denote the inputs to the VLM, $h_{2D}$ and $h_{4D}$ are the feature reconstruction heads, and $Z_{\mathcal{A}}^{t}$ is the action latent produced by the action module $\mathcal{A}$ (see Eq.~\ref{eq:action-expert}). At the same time, \(Z_{\mathcal{A}}^{t}\) is fed into the action prediction head to predict the diffusion noise:
\begin{equation}
\mathcal{L}_{action} = \mathbb{E}_{\epsilon \sim \mathcal{N}(0, I)}\!\left[ \left\| h_{action}\!\left(Z_{\mathcal{A}}^{t}\right) - \epsilon \right\|_2^{2} \right],
\end{equation}

where \(\epsilon \sim \mathcal{N}(0, I)\) denotes the forward-process noise, and \(h_{action}\) is the action prediction head. Finally, the total loss for SwiftVLA is a weighted sum of the reconstruction, action prediction, and trajectory objectives:
% \begin{equation}
% \mathcal{L}_{total}
% = \lambda_{2D}\,\mathcal{L}_{2D}
% + \lambda_{4D}\,\mathcal{L}_{4D}
% + \lambda_{action}\,\mathcal{L}_{action}
% + \lambda_{traj}\,\mathcal{L}_{traj},
% \end{equation}
\begin{equation}
\begin{split}
\mathcal{L}_{total} &= \lambda_{2D}\,\mathcal{L}_{2D} + \lambda_{4D}\,\mathcal{L}_{4D} \\
                    &\quad + \lambda_{action}\,\mathcal{L}_{action} + \lambda_{traj}\,\mathcal{L}_{traj},
\end{split}
\end{equation}
Where each \(\lambda\) serves as a balancing coefficient among the objectives. This design encourages the model to learn more comprehensive and geometry-aware 4D representations, rather than relying on a single modality for action prediction.
Meanwhile, this mechanism enables the model to implicitly reconstruct and reason over 4D spatial structures even when explicit 4D feature inputs are unavailable.
% \begin{figure}[t]
%   \centering
%   \includegraphics[width=0.99\linewidth]{figs/mask4.pdf}
%   \caption{The attention structure during training. Gray and white blocks denote visible and non-visible tokens, while orange blocks represent 4D tokens that are masked with some probability.}
%   \label{fig:mask1}
% \end{figure}

\noindent
\textbf{Inference.} To further reduce the overall parameter count and facilitate deployment on edge  platforms, we retain only the 2D feature branch during inference. In this stage, the 4D feature extractor, reconstruction heads, and trajectory head are removed, as they are only used for auxiliary supervision during training. Consequently, the deployed model consists solely of the VLM and the action expert, forming a compact yet effective architecture. The total parameter count of the deployed model equals the sum of these two components. Despite its lightweight nature, this design preserves the strong spatiotemporal perception capability learned through masked training, enabling efficient and reliable deployment on real-world robotic platforms.

% \begin{figure}[t]
%   \centering
%   \includegraphics[width=0.99\linewidth]{figs/task_example.pdf}
%   \caption{The attention structure during training. Gray and white blocks denote visible and non-visible tokens, while orange blocks represent 4D tokens that are masked with some probability.}
%   \label{fig:mask1}
% \end{figure}

% To reduce the overall parameter count and enable edge deployment, we retain only the 2D feature branch at inference. We remove the 4D feature extractor, the reconstruction heads, and the trajectory head. As a result, the deployed model consists solely of the VLM and the action expert, and its parameter count equals the sum of these two modules.

%% file: sec/4_experiments.tex
\section{Experiments}

\begin{table*}[t]
\centering
\setlength{\abovecaptionskip}{0.5em}
\setlength{\tabcolsep}{12pt}
\resizebox{0.99\linewidth}{!}{
\begin{tabular}{lccccccccc}
\toprule
\multirow{2}{*}{Methods} & \multicolumn{2}{c}{Clean the desk} & \multicolumn{2}{c}{Throw the bottle} & \multicolumn{2}{c}{Stack Bowls} & \multicolumn{2}{c}{Average} \\
\cmidrule(lr){2-3} \cmidrule(lr){4-5} \cmidrule(lr){6-7} \cmidrule(lr){8-9}
& SR $\uparrow$ & Length $\downarrow$ & SR $\uparrow$ & Length $\downarrow$ & SR $\uparrow$ & Length $\downarrow$ & SR $\uparrow$ & Length $\downarrow$ \\
\midrule
$\pi_0$~\cite{pi_0} & \secondbest{0.60} & 1220 & 0.66 & \secondbest{980} & 0.56 & 840 & 0.61 & 1013 \\
SmolVLA~\cite{smolvla} & 0.32 & 1640 & 0.40 & 1360 & 0.30 & 1360 & 0.34 & 1453 \\
SmolVLA\textsuperscript{$\dagger$}~\cite{smolvla} & 0.52 & 1360 & 0.54 & 1140 & 0.52 & 860 & 0.53 & 1120 \\
\rowcolor{mygray}
SwiftVLA & \best{0.86} & \secondbest{1140} & \secondbest{0.80} & \secondbest{980} & \secondbest{0.74} & \best{800} & \secondbest{0.80} & \secondbest{973} \\
\rowcolor{mygray}
SwiftVLA with 4D input & \best{0.86} & \best{1090} & \best{0.82} & \best{960} & \best{0.78} & \secondbest{810} & \best{0.82} & \best{953} \\
\bottomrule
\end{tabular}
}
\caption{Comparison of task success rate and average trajectory length in real-world experiments. The best results are marked in \best{bold}, and the second-best results are \secondbest{underlined}. {$\dagger$} denotes the model that is pre-trained and fine-tuned using the same configuration as SwiftVLA.}
\label{tab:real}
\vspace{-1.5mm}
\end{table*}

\begin{table}[t]
\centering
\small
\setlength{\abovecaptionskip}{0.5em}
\resizebox{0.99\linewidth}{!}{
\begin{tabular}{lcccccc}
\toprule
\multirow{2}{*}{\makecell[c]{\vspace*{-0.3cm}\\ Methods}} & \multirow{2}{*}{\makecell[c]{\vspace*{-0.3cm}\\ Size}} & \multicolumn{5}{c}{LIBERO} \\
\cmidrule(lr){3-7}
 & & Spatial & Object & Goal & Long & Avg \\
\midrule
\multicolumn{7}{c}{Spatio-Temporal Enhanced VLA} \\
\midrule
SpatialVLA~\cite{spatialvla} & 4B & 88.2 & 89.9 & 78.6 & 55.5 & 78.1 \\
4D-VLA~\cite{4DVLA} & 4B & 88.9 & 95.2 & 90.9 & 79.1 & 88.6 \\
QDepth-VLA~\cite{li2025qdepth} & 4B & \best{97.6} & 96.6 & 95.2 & 90.0 & 94.9 \\
\midrule
\multicolumn{7}{c}{Small VLA} \\
\midrule
SmolVLA~\cite{smolvla} & \best{0.45B} & 90.0 & 96.0 & 92.0 & 71.0 & 87.3 \\
SmolVLA\textsuperscript{$\dagger$}~\cite{smolvla} & \best{0.45B} & 93.5 & 96.5 & 95.4 & 83.4 & 92.2 \\
UniAct~\cite{zheng2025universal} & \secondbest{0.5B} & 77.0 & 87.0 & 77.0 & 70.0 & 77.8 \\
VLA-OS~\cite{vlaos}  & \secondbest{0.5B} & 87.0 & 96.5 & 92.7 & 66.0 & 85.6 \\
SmolVLA~\cite{smolvla} & 2B & 93.0 & 94.0 & 91.0 & 77.0 & 88.8 \\
\midrule
\multicolumn{7}{c}{Large VLA} \\
\midrule
GR00T-N1~\cite{bjorck2025gr00t} & 3B & 94.4 & 97.6 & 93.0 & 90.6 & 93.9 \\
$\pi_{0}$~\cite{pi_0} & 3B & 96.8 & \best{98.8} & 95.8 & 85.2 & 94.1 \\
$\pi_{0}$+FAST~\cite{pertsch2025fast} & 3B & 96.4 & 96.8 & 88.6 & 60.2 & 85.5 \\
OpenVLA~\cite{openvla} & 7B & 84.7 & 88.4 & 79.2 & 53.7 & 76.5 \\
OpenVLA-OFT~\cite{openvlaoft} & 7B & \best{97.6} & 98.4 & \best{97.9} & \best{94.5} & \best{97.1} \\
DD-VLA~\cite{liang2025discrete} & 7B & \secondbest{97.2} & \secondbest{98.6} & \secondbest{97.4} & 92.0 & \secondbest{96.3} \\
UniVLA~\cite{wang2025unified} & 9B & 95.4 & \best{98.8} & 93.6 & \secondbest{94.0} & 95.4 \\
\midrule
\multicolumn{7}{c}{Spatio-Temporal Enhanced Small VLA} \\
\midrule
\rowcolor{mygray}
SwiftVLA  & \best{0.45B} & 97.0 & 96.4 & 96.8 & 88.4 & 94.7\\
\rowcolor{mygray}
SwiftVLA with 4D input & 1.65B & \secondbest{97.2} & 96.8 & \secondbest{97.4} & 89.0 & 95.1 \\
\bottomrule
\end{tabular}}
\caption{Comparison of methods on the LIBERO. The best results are marked in \best{bold}, and the second-best results are \underline{underlined}. {$\dagger$} denotes the model that is pre-trained and fine-tuned using the same  configuration as SwiftVLA.}
\label{LIBERO}
\vspace{-8pt}
\end{table}

\subsection{Experimental Setup}
\noindent
\textbf{Evaluation Metrics.}
We employ success rate (SR) primarily as our evaluation metric, along with the average trajectory length. In simulations, a task receives an SR of 1 for successful completion and 0 otherwise. For real-world evaluations, we use a detailed scoring system, where in the pick-and-place task, a score of 0.5 is given for grasping the object and another 0.5 for placing it at the target location.

\noindent
\textbf{Baselines.} We primarily selected VLA models of different parameter sizes as baselines for comparison with SwiftVLA. For large models, we chose the current state-of-the-art model $\pi_0$~\cite{pi_0} and GO-1~\cite{bu2025agibot}. For smaller models, we selected TinyVLA~\cite{tinyvla} and SmolVLA~\cite{smolvla}. For SwiftVLA, we adopted two inference configurations: one that uses 4D inputs during inference, referred to as SwiftVLA with 4D input, and another that does not use 4D inputs during inference, referred to simply as SwiftVLA. Both configurations share the same set of trained weights. Additionally, in the LIBERO benchmark~\cite{liu2023libero}, we compared several other algorithms, grouped into three categories: spatio-temporal enhanced VLA models~\cite{spatialvla,4DVLA,li2025qdepth}, which utilize 3D or 4D inputs; small VLA models~\cite{smolvla,vlaos,zheng2025universal}, which employ smaller VLMs; and large VLA models~\cite{bjorck2025gr00t,pi_0,pertsch2025fast,openvla,openvlaoft,liang2025discrete,wang2025unified}, which refer to VLA models with more than 3B parameters.

\noindent
\textbf{Implementation Details.}
We adopt SmolVLM~\cite{smolvlm} as the backbone. The complete model comprises approximately 450 million parameters, of which around 100 million are allocated to the action-expert module. Meanwhile, we pretrain our model on public datasets~\cite{bu2025agibot,wu2024robomind} using a two-stage training procedure (detailed in the appendix).
% We adopt SmolVLM-2~\cite{smolvlm} as the backbone. The complete model comprises 450 million parameters, of which approximately 10 0 million are allocated to the action-expert module. Meanwhile, we pretrain our model on public datasets~\cite{bu2025agibot,robotwin,wu2024robomind} using a two-stage procedure (detailed in the appendix).
\subsection{Simulation Benchmark Experiments}

\noindent
\textbf{Simulation Setup.} We conduct a systematic evaluation of SwiftVLA on both the RoboTwin 2.0~\cite{robotwin} and LIBERO benchmarks~\cite{liu2023libero}. For RoboTwin 2.0, our experimental setup considers three categories of tasks: Short-Horizon, Medium-Horizon, and Long-Horizon, with two subtasks selected for each category. For each subtask, we generate 50 demonstration trajectories, which are then used for post-training. For LIBERO benchmark, we perform experiments across four task suites: LIBERO-Spatial, LIBERO-Object, LIBERO-Goal, and LIBERO-Long.

% As shown in Fig.~\ref{simulation}, we illustrate the tasks in the simulation. More details are provided in the appendix.

\noindent
\textbf{Results.}
As shown in Tab.~\ref{tab:robotwin_results}, the mask-and-reconstruct strategy enables SwiftVLA, which does not use 4D input during inference, to maintain competitiveness comparable to that of SwiftVLA with 4D input. Moreover, SwiftVLA demonstrates strong performance across all three task categories, exhibiting competitive capabilities compared to \(\pi_0\)~\cite{pi_0} while using only about 15\% of its parameters. Meanwhile, TinyVLA~\cite{tinyvla} and SmolVLA~\cite{smolvla} still lag notably behind $\pi_0$~\cite{pi_0}, mainly due to their smaller VLM backbones, which lack sufficient capacity to model long-term spatiotemporal dependencies. In contrast, SwiftVLA introduces 4D representations to enhance its spatiotemporal understanding, improving SR by 82.76\% over SmolVLA~\cite{smolvla}.

As shown in Tab.~\ref{LIBERO}, we compare SwiftVLA with three categories of methods on LIBERO. For spatio-temporal enhanced VLA~\cite{spatialvla,4DVLA,li2025qdepth}, methods like SpatialVLA~\cite{spatialvla} and 4D-VLA~\cite{4DVLA} directly integrate 3D and 4D information into the VLM. However, their performance gains are often limited by the model's ability to handle multiple modalities. Approaches such as Qdepth-VLA~\cite{li2025qdepth} add additional 3D processing branches, improving performance but increasing the model size to over 3B parameters. Smaller models~\cite{smolvla,vlaos}, despite having fewer than 2B parameters, typically show lower success rates. Large VLA models offer strong performance but also require over 3B parameters, making them costly for real-world deployment. In contrast, SwiftVLA effectively leverages 4D information while maintaining a compact design, achieving performance comparable to large VLA models.

\subsection{Real-World Experiment}
\begin{figure}[t]
    \centering
    \includegraphics[width=.99\linewidth]{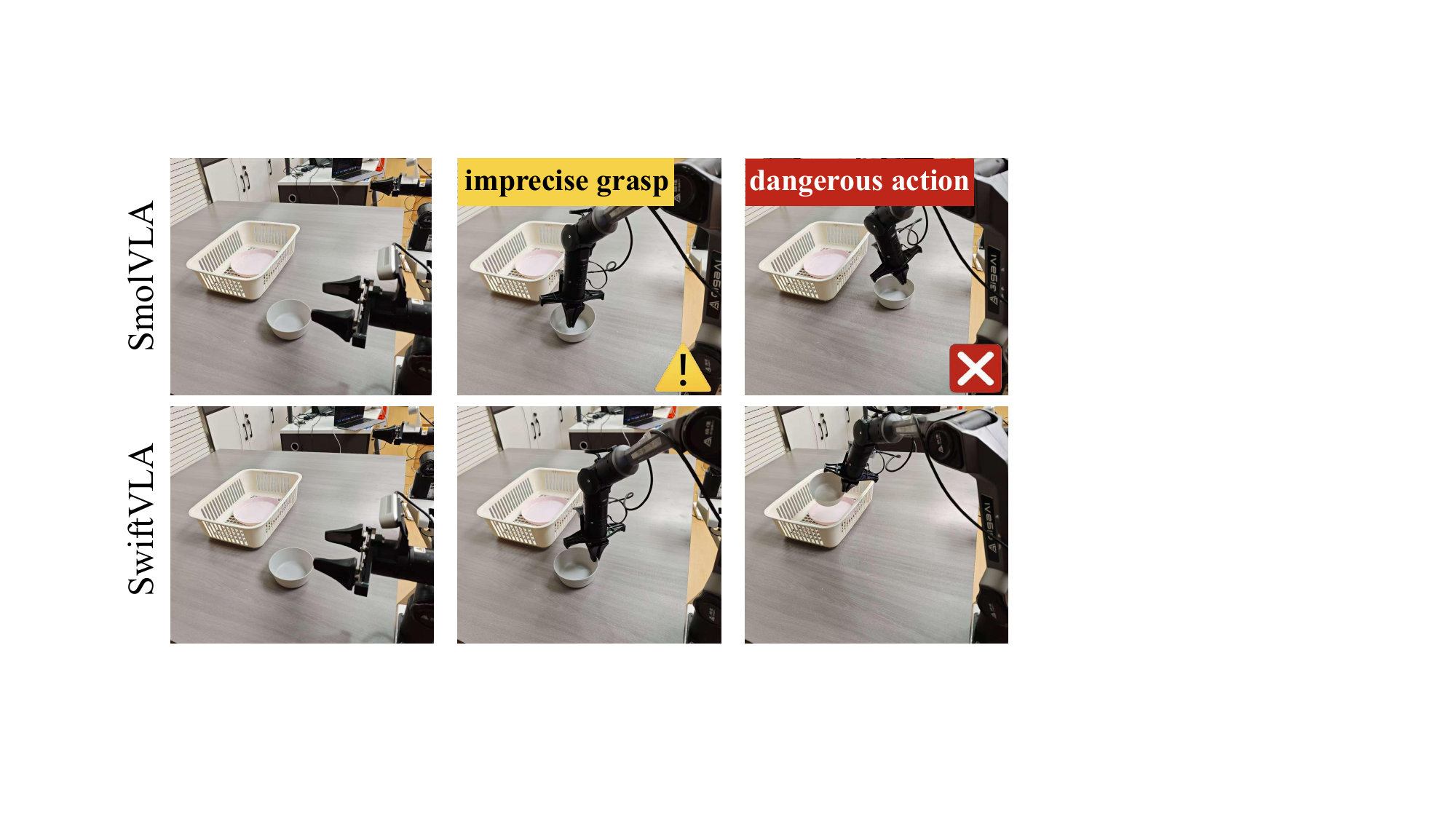} 
    % \vspace{-2mm}
    \caption{Comparison of SmolVLA and SwiftVLA under identical initial poses. During execution, SmolVLA fails to grasp accurately, as the end-effector misses the target and collides with the object, causing it to shift and posing safety risks. In contrast, SwiftVLA successfully completed the grasp with accurate positioning and stable control, demonstrating superior performance.}
    % \caption{Comparison of SmolVLA and SwiftVLA under identical initial poses. During execution, SmolVLA failed to achieve precise grasping, as the end-effector did not accurately reach the target position and collided with the object, causing it to shift and potentially leading to safety risks. In contrast, SwiftVLA successfully completed the grasp with accurate positioning and stable control, demonstrating superior performance.}
% \caption{We present a comparison between SmolVLA and SwiftVLA under identical initial placements. During execution, SmolVLA fails to perform precise grasping, causing the end-effector to collide with the target object, resulting in a shift in the object’s position, as indicated by the red arrow. More critically, after grasping the object, an incorrect downward force is applied, causing it to tilt. This compromises the stability of the robotic arm, potentially leading to grasp failure or the object falling. In contrast, SwiftVLA successfully completes the task, accurately grasping the object and avoiding the aforementioned issues.
% }
    \label{realexp}
    \vspace{-1mm}
\end{figure}

To evaluate the effectiveness of the method in the real world, we conducted a gripper grasping experiment using the AgileX PiPER six-degree-of-freedom robotic arm, with computational support provided by an NVIDIA RTX 4090 GPU. In addition, we designed a series of real-world tasks, including Clean the Desk, Throw the Bottle, and Stack Bowls, with detailed descriptions provided in the supplementary material. As summarized in Tab.~\ref{tab:real}, our method demonstrates strong performance compared to $\pi_0$~\cite{pi_0} while using fewer parameters, and significantly outperforms similarly sized baselines such as SmolVLA. As shown in Fig.~\ref{realexp}, we compared SmolVLA and SwiftVLA under identical initial object placements. During execution, SmolVLA failed to achieve precise grasping due to its limited understanding of geometric information. Meanwhile, the end-effector collided with the target object and displaced it, which could lead to task failure or safety hazards. In contrast, SwiftVLA successfully performed stable and accurate grasping thanks to its superior spatial perception and control.

% To evaluate real-world effectiveness, we conducted gripper–grasp experiments using an AgileX PiPER 6-DoF robotic arm. The sensing suite consisted of three Intel RealSense D435 cameras: a static, front-facing camera observing the workspace, and two wrist-mounted cameras positioned on the left and right sides of the end-effector. In each episode, the robot is permitted up to 20 consecutive grasp attempts while objects are randomly arranged on the tabletop. As summarized in Tab.~\ref{tab:real}, our method achieves a success rate comparable to that of $\pi_0$~\cite{pi_0}, while utilizing fewer parameters, and it substantially outperforms equally sized baselines, such as SmolVLA~\cite{smolvla}, highlighting the importance of incorporating 4D information.

\begin{table}[t]
\centering
\resizebox{0.99\linewidth}{!}{
\begin{tabular}{lcccc}
\toprule
Methods &  Inference Time (s) & Memory (MB)  & Average SR \\
\midrule
 $\pi_0$~\cite{pi_0}  &  2.966 & 16236.2  & 0.48 \\
SmolVLA~\cite{smolvla}  & 0.166 & 1397.5  & 0.30 \\
\rowcolor{mygray}
SwiftVLA    & 0.167 & 1398.4  & 0.76 \\
\bottomrule
\end{tabular}
}
\vspace{-1.5mm}
\caption{Comparison of inference time, memory usage, and SR across models on the NVIDIA Jetson Orin~\cite{nvidia_jetson_orin}.}
\label{tab:inference_time}
\vspace{-3mm}
\end{table}

\subsection{Deployment on the Edge Device}
To evaluate the deployment efficiency of VLAs on edge devices, we adopt the NVIDIA Jetson Orin~\cite{nvidia_jetson_orin} as our target platform. As summarized in Tab.~\ref{tab:inference_time}, we report the inference time and parameter count of each model. The results show that SwiftVLA delivers an \(18\times\) speedup in inference compared to the larger VLA model \(\pi_0\), while reducing memory consumption by \(12\times\). Moreover, its inference latency is nearly identical to that of SmolVLA. Meanwhile, SwiftVLA demonstrates a high success rate compared to $\pi_0$ on the NVIDIA Jetson Orin~\cite{nvidia_jetson_orin}, indicating its suitability.

\begin{table}[t]
\centering
\setlength{\tabcolsep}{17pt} % 控制列间距
\setlength{\abovecaptionskip}{0.5em}
\resizebox{0.99\linewidth}{!}{
\begin{tabular}{ccc}
\toprule
\makecell[c]{Input Feature Type} & 
\textit{Fusion tokens} & 
Average SR\\
\midrule
2D &      & 0.36 \\
2D \& 4D &         & \secondbest{0.40} \\
2D \& 4D  &   $ \checkmark$       & \best{0.50} \\
\bottomrule
\end{tabular}
}
\caption{Ablation study of SwiftVLA with 4D input evaluated by SR. The input feature type indicates the modalities used during training and inference. The best results are marked in \best{bold}, and the second-best results are \underline{underlined}.}
\label{tab:ablation}
\vspace{-3.5mm}
\end{table}

\subsection{Ablation Study}

In this section, we perform experiments on the RoboTwin 2.0~\cite{robotwin} platform to address the following questions.

\noindent
\textbf{Q1. How do 4D features affect task success?} We compare two settings that use 2D inputs and that combine 2D inputs with the 4D features, as shown in the first and second rows of Tab.~\ref{tab:ablation}. The results indicate that relying solely on 2D inputs leads to a lower SR. Incorporating the 4D features yields a substantial improvement, suggesting that 4D features provide stronger representations for action planning.

\noindent
\textbf{Q2: What role do \textit{Fusion Tokens} play?} \textit{Fusion Tokens} are designed to integrate 4D and 2D features, using the 2D end-effector trajectory as supervision for trajectory prediction. In the second and third rows of Tab.~\ref{tab:ablation}, we compare models with and without \textit{Fusion Tokens} and observe significant improvements when the token is enabled. This is because small models struggle to fully leverage the input 4D information. The introduction of \textit{Fusion Tokens}, along with the design of a target task, helps guide the model to effectively use both 2D and 4D cues, leading to improved cross-modal alignment and more effective temporal cue utilization.

% \textit{Fusion Tokens} are designed to fuse 4D and 2D features, using the 2D end‑effector trajectory as supervision for trajectory prediction. In the second and third rows of Tab.~\ref{tab:ablation}, we compare models with and without \textit{Fusion Tokens} and observe significant gains when the token is enabled, as the smll models hard to 充分利用输入的4D信息， fusion tokens的引入和一个目标任务的设计，引导了模型利用2D和4D   信息， improved cross-modal alignment and more effective utilization of temporal cues.

% We adopt a mask-and-reconstruction strategy that randomly drops 2D or 4D features with a fixed probability. During training, the VLA predicts actions while reconstructing the masked features. The goal of this strategy is to enable the model to maintain performance comparable to that with full 4D input, even when 4D information is absent during inference.  

\noindent
\textbf{Q3. What is the effect of the mask-and-reconstruction strategy?} We employ a mask-and-reconstruction strategy during training, where 2D or 4D features are randomly dropped with a fixed probability, and the VLA is tasked with reconstructing the masked features. The aim is to enable the model to maintain performance comparable to full 4D input, even when 4D information is missing during inference. As shown in Tab.~\ref{tab:mask}, we compare different training strategies and evaluate performance under both inference with and without 4D input. The results show that directly removing the 4D input during inference, without applying any strategy, leads to a significant performance drop, as the model becomes overly dependent on 4D cues for prediction. Introducing 4D feature masking alleviates this dependency and preserves part of the performance when 4D input is unavailable. Moreover, incorporating the feature reconstruction helps distill 4D information into the VLA during training, allowing the model to achieve performance comparable to that with full 4D input, even in the absence of 4D features during inference. In addition, we find that moderately masking 2D features encourages the model to better exploit underlying 4D geometric cues and enhances cross-modal consistency, as reflected in the last column of Tab.~\ref{tab:mask}.

\noindent
\textbf{Q4. How does the cached memory size \(K\) affect performance?} We analyze how the cached memory size \(K\) chosen during training affects model performance. We evaluate four fixed settings with \(K \in \{3,4,5,6\}\) and a randomized strategy that samples \(K \in \{3,4,5,6\}\) at each training step.  As shown in Tab.~\ref{tab:cache}, the randomized strategy outperforms all fixed-length baselines, indicating that exposure to variable temporal horizons process notably enhances adaptability.

\begin{table}[t]
\centering
\setlength{\tabcolsep}{4pt} % 控制列间距
\resizebox{0.99\linewidth}{!}{
\begin{tabular}{ccccc}
\toprule
\thead{4D Feature \\ Mask} &
\thead{2D Feature \\ Mask} &
\thead{Feature \\ Reconstruction} &
\thead{SwiftVLA} &
\thead{SwiftVLA with \\ 4D Input} \\
\midrule
 &  &  & 0.02 & 0.50 \\
$\checkmark$ &  &  & 0.40 & 0.48 \\
$\checkmark$ &  & $\checkmark$ & \secondbest{0.50} & \secondbest{0.52} \\
$\checkmark$ & $\checkmark$ & $\checkmark$ & \best{0.53} & \best{0.55} \\
\bottomrule
\end{tabular}}
\vspace{-1.7mm}
\caption{Comparison of SR across different training strategies. All models are trained with both 2D and 4D feature inputs, while only 2D features are used during inference. The best results are marked in \best{bold}, and the second-best results are \underline{underlined}.}
\label{tab:mask}
% \vspace{-2mm}
\end{table}

\begin{table}[t]
\centering
\small
\begin{tabular}{lcc}
\toprule
Size & SwiftVLA & SwiftVLA with 4D input \\
\midrule
$K{=}3$ & 0.47 & 0.49 \\
$K{=}4$ & 0.48 & \secondbest{0.52} \\
$K{=}5$ & 0.50 & 0.51 \\
$K{=}6$ & \secondbest{0.52} & \best{0.55} \\
Random & \best{0.53} & \best{0.55} \\ 
\bottomrule
\end{tabular}
\vspace{-1.7mm}
\caption{Ablation study on the temporal cache size evaluated by SR. All models are trained with both 2D and 4D feature inputs, while only 2D features are used during inference. The best results are marked in \best{bold}, and the second-best results are \underline{underlined}.}
\vspace{-5mm}
\label{tab:cache}
\end{table}

%% file: sec/5_conclusion.tex
\section{Conclusion}
In this paper, we present SwiftVLA, a lightweight framework that achieves strong spatiotemporal reasoning while maintaining design efficiency. Specifically, we employ a 4D visual geometry transformer with a temporal cache that extracts 4D features and integrates them into the VLM to enhance both spatial and temporal modeling. To bridge the gap between 2D and 4D features, we introduce \textit{Fusion Tokens}, whose representations are supervised by the future trajectory of the end-effector, effectively capturing integrated multimodal information. Additionally, we employ a mask-and-reconstruct strategy to distill 4D knowledge into the VLA, while minimizing performance degradation when 4D inputs are omitted during inference. Experiments show that SwiftVLA matches the performance of models with up to \(7 \times\) more parameters, while offering up to \(18 \times\) faster inference and \(12 \times\) smaller memory footprint on edge devices.

%% file: sec/X_suppl.tex
\appendix
% \setcounter{page}{1}
% \maketitlesupplementary

\newpage
   \twocolumn[
    \centering
    \Large
    \textbf{SwiftVLA: Unlocking Spatiotemporal Dynamics for Lightweight VLA\\ Models at Minimal Overhead}\\
    \vspace{0.5em}Supplementary Material \\
    \vspace{1.0em}
    \resizebox{0.99\linewidth}{!}{
\includegraphics{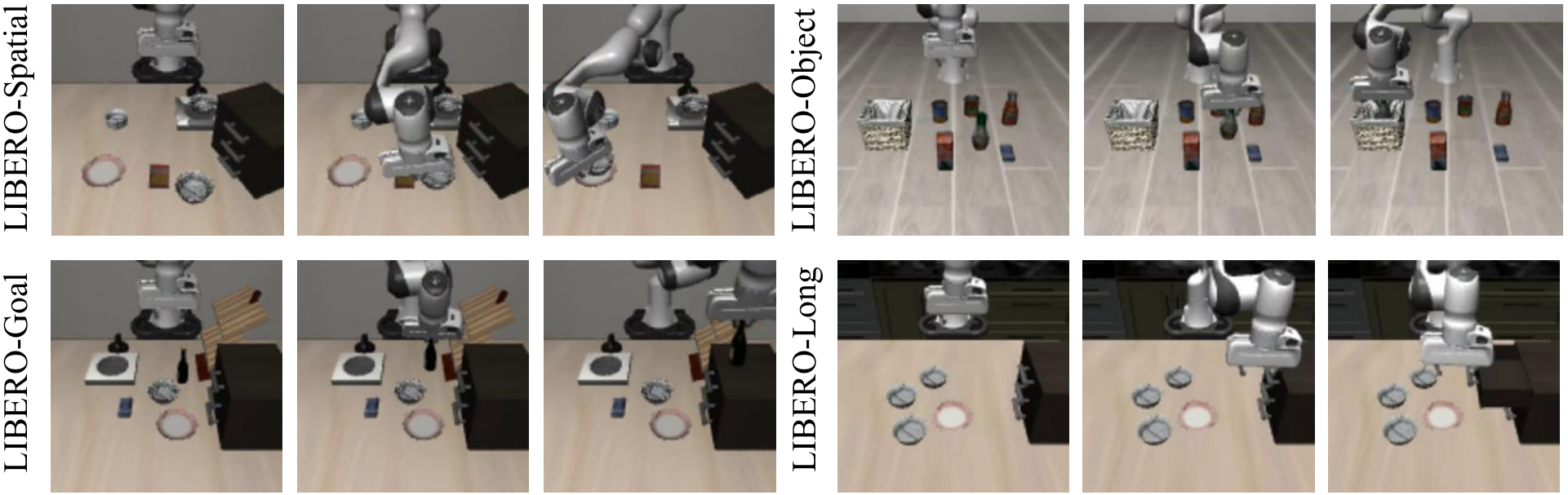
}}
\captionof{figure}{
Examples from the LIBERO-Spatial, LIBERO-Object, LIBERO-Goal, and LIBERO-Long task suites~\cite{liu2023libero}.
}
\label{fig:simulation1}
 \vspace{1.0em}
   ] %< twocolumn

% \twocolumn[
% {
% \begin{center}
% \centering
% \setlength{\abovecaptionskip}{0.5em}
% \resizebox{0.99\linewidth}{!}{
% \includegraphics{supp/lebero.pdf
% }}
% \captionof{figure}{
% We compare our method with WonderWorld~\citep{wonderworld} under novel views. The black-box regions exhibit noticeable artifacts in WonderWorld~\citep{wonderworld}, whereas WonderFree maintains clarity and visual consistency.
% }
% \label{fig:teaser}
% \end{center}}]

% \begin{figure*}[t]
%     \centering
%     \includegraphics[width=.99\linewidth]{supp/lebero.pdf} 
%     % \vspace{-2mm}
%     \caption{Comparison of SmolVLA and SwiftVLA under identical initial poses. During execution, SmolVLA fails to grasp accurately, as the end-effector misses the target and collides with the object, causing it to shift and posing safety risks. In contrast, SwiftVLA successfully completed the grasp with accurate positioning and stable control, demonstrating superior performance.*}
%     \label{realexp}
%     \vspace{-1mm}
% \end{figure*}

\begin{figure*}[t]
    \centering
    \includegraphics[width=.99\linewidth]{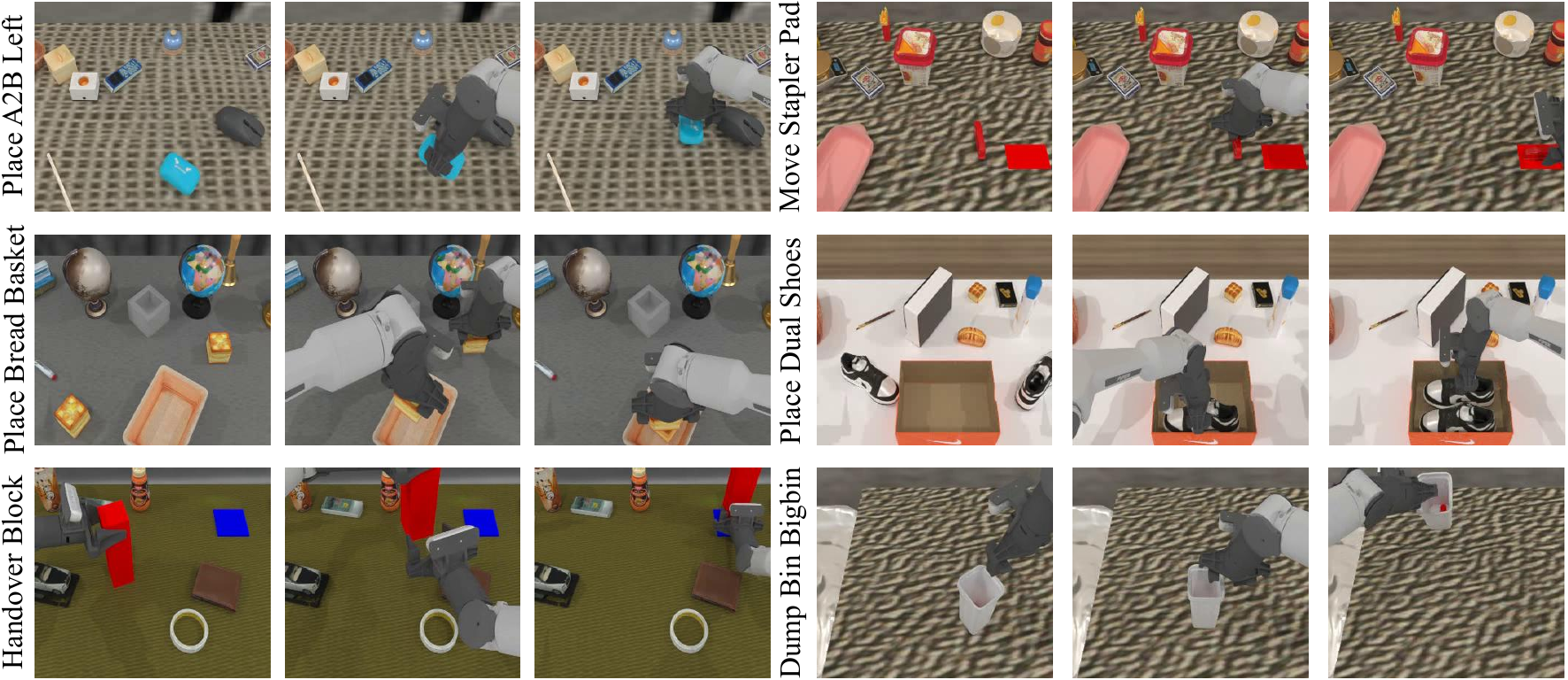} 
    % \vspace{-2mm}
    \caption{Examples from the RoboTwin 2.0~\cite{robotwin}, including move Stapler Pad, Place A2B Left, Place Bread Basket, Place Dual Shoes, Dump Bin Bigbin, Handover Block.}
    \label{simulation2}
    \vspace{-2mm}
\end{figure*}

\section{Architecture Design.}
We adopt SmolVLM~\cite{smolvlm} as the backbone network for robot environmental perception. SmolVLM~\cite{smolvlm} utilizes SigLIP~\cite{zhai2023sigmoid} to encode visual features, which are then passed to the language decoder. Additionally, by leveraging global image information and employing pixel-shuffle operations, we constrain the number of visual tokens per frame to 64. To further accelerate inference, we skip certain computational layers in the VLM by using only the first 16 layers of the model.
To enhance both computational efficiency and response speed, we adopt an attention pattern that alternates between self-attention~\cite{pi_0} and cross-attention~\cite{bjorck2025gr00t} modules rather than relying solely on either mechanism, following the design principles of SmolVLA~\cite{smolvla}. For handling 4D features, we similarly limit the number of tokens to 64, ensuring consistent computational efficiency.

\section{Implementation Details.}

\noindent
\textbf{Baselines.}
We primarily compare our model with VLA models of different parameter scales, using them as baselines for evaluation.
 
$\pi_0$~\cite{pi_0} is a VLM~\cite{paligemma} that incorporates Flow Matching~\cite{lipman2022flow} to predict action chunks. With a parameter count of 3.3 billion, it has been trained on a dataset comprising 10,000 hours of cross-embodiment robotics data. The architecture is inspired by Paligemma~\cite{paligemma} and processes three images, sensorimotor states, and a language instruction as inputs.

TinyVLA~\cite{tinyvla} is designed to address the challenges of inference speed and data efficiency in existing VLA models. Unlike traditional models, TinyVLA~\cite{tinyvla} achieves faster inference and improved data efficiency by initializing a high-performance multimodal policy backbone and incorporating a diffusion policy decoder during finetuning. With a model size around 1 billion parameters, TinyVLA~\cite{tinyvla} demonstrates advantages in both speed and data utilization,

SmolVLA~\cite{smolvla} is a compact and efficient VLA model designed to reduce training and inference costs, making it suitable for real-world robotics applications. Optimized for consumer-grade GPUs, it retains competitive performance despite its small size. The model is pre-trained on community-collected datasets with fewer than 30k episodes and features an asynchronous inference stack for faster and more responsive control. SmolVLA~\cite{smolvla} performs on par with larger models, offering a solution for robotics tasks in both simulated and real-world environments.

\noindent
\textbf{Pretraining Details.} We pretrain our model on public datasets~\citep{bu2025agibot,wu2024robomind} using a two-stage procedure. In the first stage, the model is trained without 4D inputs, \textit{Fusion Tokens}, or the mask-and-reconstruct strategy, relying solely on robot actions for supervision. Training is performed for 100{,}000 steps with a global batch size of 256. The learning rate follows a cosine decay schedule, starting at $1\times10^{-4}$ and decaying to $2.5\times10^{-6}$ after a 200-step warm-up. We adopt the AdamW optimizer~\citep{kingma2014adam} with $\beta_{1}=0.85$ and $\beta_{2}=0.9$. The input images are resized to $512\times512$ pixels for compatibility with the vision-language encoder. In the second stage, the model is initialized from the first-stage checkpoint, and 4D inputs, \textit{Fusion Tokens}, are enabled along with the mask-and-reconstruct strategy. Training continues for an additional 50{,}000 steps under the same optimizer settings, with a reduced learning rate of $5\times10^{-5}$ following a cosine decay schedule.

\noindent
\textbf{Finetuning Details.}
For all baseline methods, we train each model for 30{,}000 steps on the same dataset, keeping hyperparameters consistent with their original implementations to ensure a fair comparison.  For SwiftVLA, we also adopt a two-stage finetuning strategy. In the first stage (the initial 10{,}000 steps), the model is supervised only with robot actions to stabilize adaptation within the action space. The learning rate follows a cosine decay schedule and is initialized at $1\times10^{-4}$. We use the AdamW optimizer~\citep{kingma2014adam} with $\beta_{1}=0.85$ and $\beta_{2}=0.9$.  After completing the first stage, we enable the 4D inputs and \textit{Fusion Tokens}, and incorporate the mask-and-reconstruct strategy in the second stage. This allows the model to further learn spatiotemporal feature fusion and higher-level structural understanding.

\begin{table}[t]
  \centering
  \resizebox{1.\linewidth}{!}{%
    \begin{tabular}{lcc}
      \toprule
      Category & Task & Steps \\
      \midrule
      \multirow{2}{*}{Short-Horizon}  & Move Stapler Pad          & 112 \\
                                      & Place A2B Left  & 113 \\
      \midrule
      \multirow{2}{*}{Medium-Horizon} & Place Bread Basket      & 151 \\
                                      & Place Dual Shoes     & 155 \\
      \midrule
      \multirow{2}{*}{Long-Horizon}   & Dump Bin Bigbin     & 283 \\
                                      & Handover Block    & 313 \\
      \bottomrule
    \end{tabular}
  }
  % \vspace{-3mm}
  \caption{Tasks and their step lengths across different horizon categories used in the RoboTwin 2.0~\cite{robotwin}.}
    \label{tab:task_horizon_analysis}
     \vspace{-3mm}
\end{table}

\begin{figure*}[t]
    \centering
    \includegraphics[width=.99\linewidth]{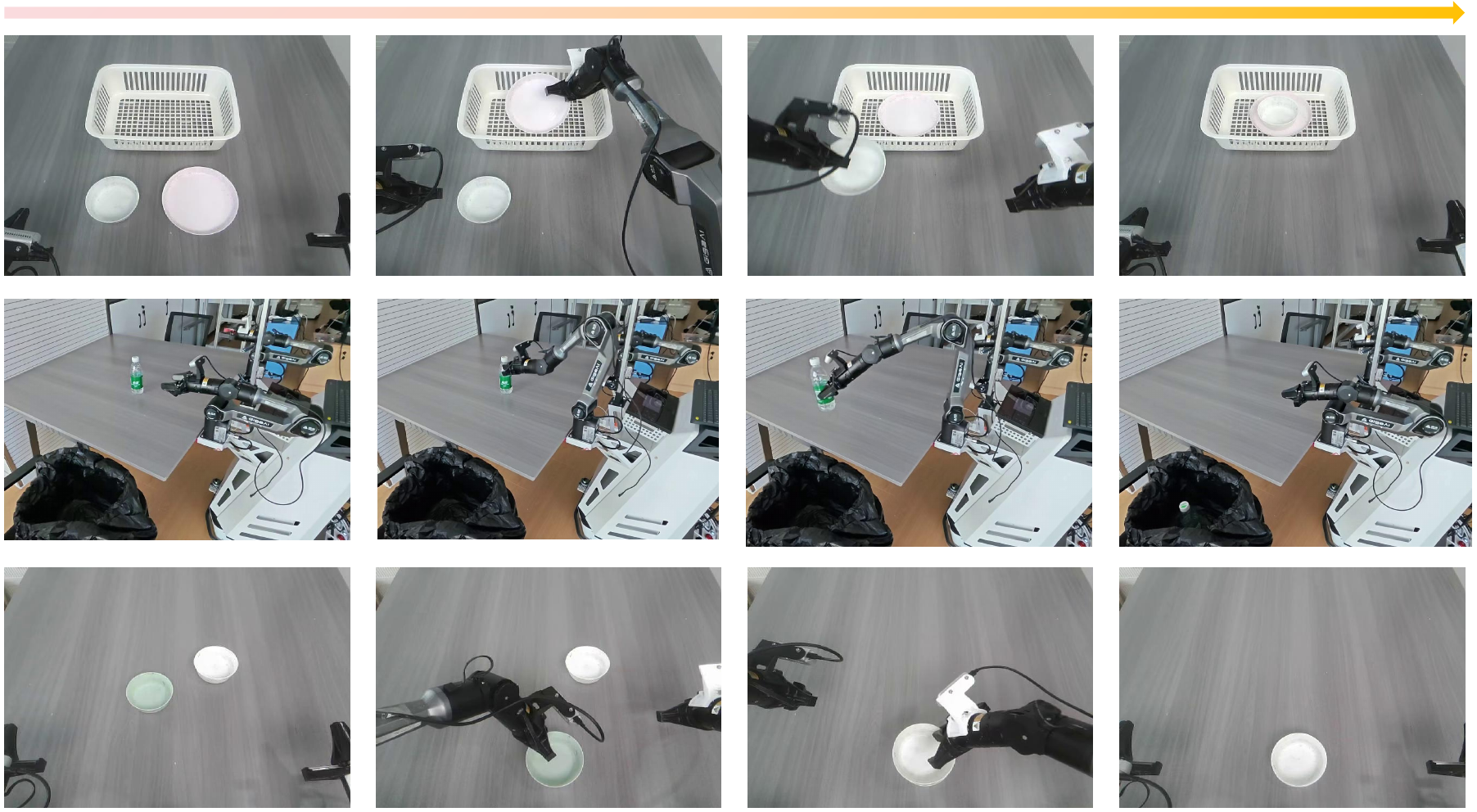} 
    % \vspace{-2mm}
    \caption{Real-world manipulation tasks used in our experiments. From top to bottom, the examples correspond to Clean the Desk, Throw the Bottle, and Stack Bowls.}
    \label{realtask}
    \vspace{-3mm}
\end{figure*}

\noindent
\textbf{Simulation Tasks Setup.}
As shown in Fig.~\ref{fig:simulation1} and Fig.~\ref{simulation2}, we present the simulation tasks from LIBERO and RoboTwin 2.0~\cite{robotwin}. In RoboTwin 2.0, we further categorize the evaluation tasks into short, medium, and long horizons based on the average number of steps required for completion. Tab.~\ref{tab:task_horizon_analysis} provides the detailed categorization. Short-horizon tasks typically require fewer than 120 steps and rely primarily on localized spatial reasoning. Medium-horizon tasks average around 150 steps and involve sequential planning across multiple object interactions. Long-horizon tasks exceed 280 steps and exhibit higher temporal dependencies and compositional complexity. This categorization enables a systematic analysis of how different models generalize across varying horizon lengths, which is crucial for assessing robustness in multi-step manipulation scenarios.

\noindent
\textbf{Real-World Tasks Setup.}
As shown in Fig.~\ref{realtask} and Fig.~\ref{cloth}, we illustrate the tasks used in our experiments, covering four manipulation tasks. 

Clean the Desk: Bowls and plates with randomized colors are placed on the table. The robot must place both items into a basket while ensuring that the plate is positioned at the bottom.

Throw the Bottle: A plastic bottle with a randomly varying amount of liquid is placed in the scene, and the robot is required to pick it up and throw it into a trash bin.

Stack Bowls: Two bowls are positioned randomly on the table, and the robot is required to stack them correctly.

Fold the Cloth: A piece of clothing is laid flat on the table. The robot folds it following a predefined sequence and then moves the folded garment to a designated location.

% As shown in Fig.~\ref{1}, 我们展示了在真机环境中的场景和任务，包括三种不同的任务。下面是对我们任务的具体描述，以及设计 clean the desk： 随机颜色的碗和盘随机进行摆放，同时随机放置一些障碍物，要求将碗和盘子准确放入篮子，并且盘子在下。throw the bottle 随机放置塑料瓶，要求拿起塑料瓶扔入垃圾桶。stack bowls 随机摆放bowl的位置，把两个bowl进行堆叠。叠衣服，将平铺的衣服进行xxx 然后xxx 最后挪动到桌子指定位置。

\begin{table}[t]
\centering
\setlength{\abovecaptionskip}{0.5em}
\setlength{\tabcolsep}{12pt}
\resizebox{0.99\linewidth}{!}{
\begin{tabular}{lcc}
\toprule
\multirow{2}{*}{Methods} & \multicolumn{2}{c}{Fold the Cloth} \\
\cmidrule(lr){2-3}
& SR $\uparrow$ & Length $\downarrow$ \\
\midrule
$\pi_0$~\cite{pi_0} & 0.45 & 2550 \\
SmolVLA~\cite{smolvla} & 0.05 &  3200 \\
SmolVLA\textsuperscript{$\dagger$}~\cite{smolvla} & 0.30 & 2600 \\
\rowcolor{mygray}
SwiftVLA & \secondbest{0.60} & \secondbest{2100} \\
\rowcolor{mygray}
SwiftVLA with 4D input & \best{0.65} & \best{2010} \\
\bottomrule
\end{tabular}
}
\caption{Comparison of task success rate and trajectory length for “Fold the Cloth". The best results are marked in \best{bold}, and the second-best results are \secondbest{underlined}. {$\dagger$} denotes the model that is pre-trained and fine-tuned using the same configuration as SwiftVLA.}
\label{tab:cloth_result}
% \vspace{-1.5mm}
\end{table}

\noindent
\section{More Challenging Real-World Experimental Results.}
\begin{figure*}[t]
    \centering
    \includegraphics[width=.99\linewidth]{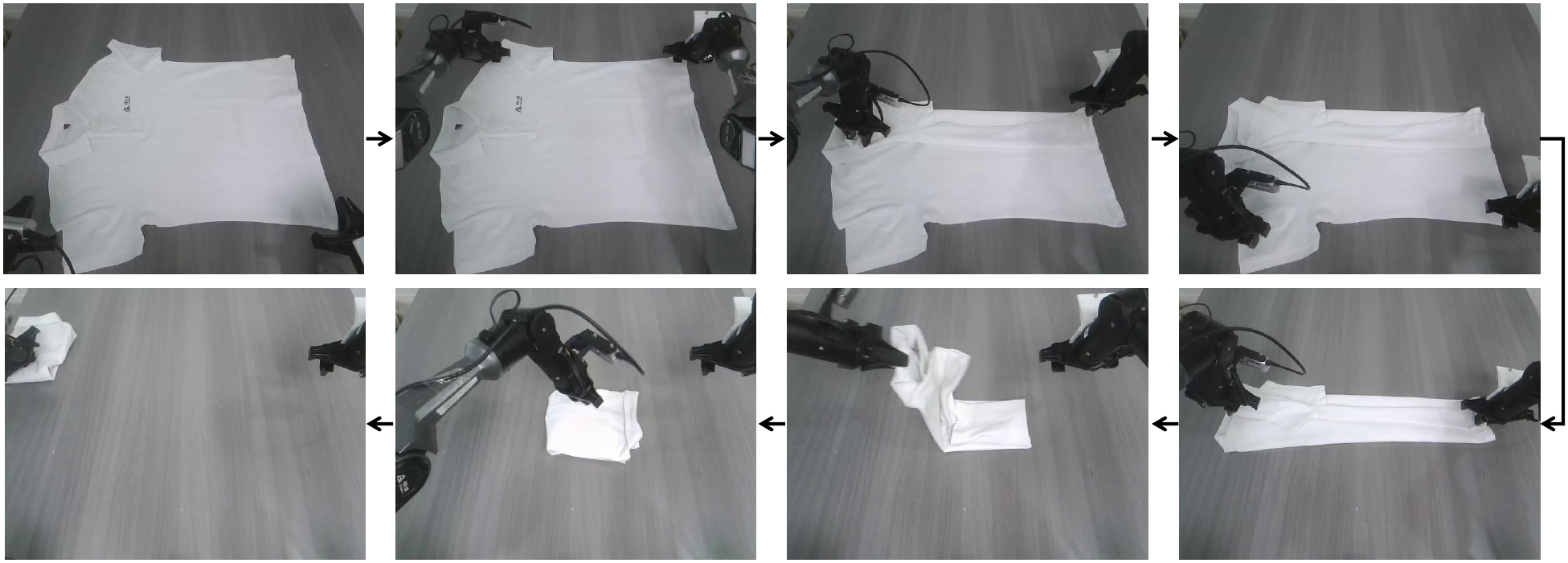} 
    % \vspace{-2mm}
    \caption{Real-world execution process of the Fold the Cloth task, which requires long-horizon reasoning and precise manipulation of deformable objects.}
    \label{cloth}
    \vspace{-3mm}
\end{figure*}

To further evaluate the real-world performance of SwiftVLA, we investigate a more challenging manipulation task: Fold the Cloth. This task is difficult due to its long-horizon nature and the complex physical dynamics of deformable objects. As shown in Fig.~\ref{cloth}, we illustrate the full execution process of this task in a real-world setting.

The results in Tab.~\ref{tab:cloth_result} present a comparison of success rates achieved by different methods on the cloth folding task, executed on the AgileX PiPER six-degree-of-freedom robotic arm with computational support provided by an NVIDIA RTX 4090 GPU. SwiftVLA demonstrates strong and reliable performance, while similar models such as SmolVLA~\cite{smolvla} achieve very low success rates. These results highlight the advantages of incorporating 4D features when handling deformable objects and long-horizon manipulation tasks. 

% To further evaluate the real-world performance of SwiftVLA, we investigate a more challenging manipulation task: Fold the Cloth. This task is difficult due to its long-horizon nature and the complex physical dynamics of deformable objects. As shown in Fig.~\ref{cloth}, we illustrate the full execution process of this task in a real-world setting.

% The results in Tab.~\ref{tab:cloth_result} present a comparison of the success rates achieved by different methods on the cloth-folding task. SwiftVLA demonstrates strong and reliable performance, while models of similar size such as SmolVLA~\cite{smolvla} achieve a success rate close to zero. These results highlight the advantages of incorporating 4D features when handling deformable objects and long-horizon manipulation tasks.

\noindent
% \section{Supplementary Video}
% We provide a video that showcases a comparison between SwiftVLA and $\pi_0$~\cite{pi_0} across multiple tasks. Please refer to the file located at video/comparison.mp4 for more details. The video consists of the following segments:

% \begin{itemize}
%     \item \textbf{4-40s:} Demonstrates the comparison between SwiftVLA and $\pi_0$~\cite{pi_0} on the “Fold the Cloth" task using an NVIDIA Jetson Orin platform~\cite{nvidia_jetson_orin}.
%     \item \textbf{40-60s:} Displays the comparison between SwiftVLA and $\pi_0$~\cite{pi_0} on the “Throw the Bottle" task using an NVIDIA Jetson Orin platform~\cite{nvidia_jetson_orin}.
%     \item \textbf{60-72s:} Compares SwiftVLA and $\pi_0$~\cite{pi_0} on the “Clean the Desk" task using an NVIDIA Jetson Orin~\cite{nvidia_jetson_orin}.
%     \item \textbf{72-90s:} Highlights the superior error-correction capability of SwiftVLA over $\pi_0$ is particularly evident when handling deformable objects. In the video, we compare the two algorithms on the "Fold the Cloth" task using an NVIDIA Jetson Orin platform~\cite{nvidia_jetson_orin} focusing on how each model adjusts after failure.  Compared to $\pi_0$, SwiftVLA is able to recover more quickly, with smoother motion trajectories that allow for more fluid and accurate handling of deformable objects.
%     \item \textbf{90-132s:} Shows additional examples of SwiftVLA on the “Fold the Cloth" task using an NVIDIA Jetson Orin~\cite{nvidia_jetson_orin}.
% \end{itemize}

\section{Supplementary Video}
We provide a video that compares SwiftVLA and $\pi_0$~\cite{pi_0} across multiple tasks. Please refer to the file located at video/comparison.mp4 for more details. The video consists of the following segments:

\begin{itemize}
    \item \textbf{4-40s:} Demonstrates the comparison between SwiftVLA and $\pi_0$~\cite{pi_0} on the “Fold the Cloth" task using an NVIDIA Jetson Orin platform~\cite{nvidia_jetson_orin}.
    \item \textbf{40-60s:} Displays the comparison between SwiftVLA and $\pi_0$~\cite{pi_0} on the “Throw the Bottle" task using an NVIDIA Jetson Orin platform~\cite{nvidia_jetson_orin}.
    \item \textbf{60-72s:} Compares SwiftVLA and $\pi_0$~\cite{pi_0} on the “Clean the Desk" task using an NVIDIA Jetson Orin~\cite{nvidia_jetson_orin}.
    \item \textbf{72-90s:} Highlights the superior error-correction capability of SwiftVLA over $\pi_0$ is particularly evident when handling deformable objects. In the video, we compare the two algorithms on the “Fold the Cloth" task using an NVIDIA Jetson Orin platform~\cite{nvidia_jetson_orin}, focusing on how each model adjusts after failure. Compared to $\pi_0$, SwiftVLA recovers more quickly, with smoother motion trajectories that enable more fluid and accurate handling of deformable objects.
    \item \textbf{90-132s:} Shows additional examples of SwiftVLA on the “Fold the Cloth" task using an NVIDIA Jetson Orin~\cite{nvidia_jetson_orin}.
\end{itemize}